\documentclass[conference]{IEEEtran}
\IEEEoverridecommandlockouts
\usepackage{cite}
\usepackage{amsmath,amssymb,amsfonts}
\usepackage{graphicx}
\usepackage{textcomp}
\usepackage{xcolor}
\usepackage{booktabs}
\usepackage{stfloats}
\usepackage{algpseudocode} 
\usepackage{multirow}
\usepackage[hidelinks]{hyperref}
\usepackage{eso-pic}
\usepackage{url}

\begin{document}

\IEEEpubid{\begin{minipage}{\textwidth}\ \\[12pt] \centering
  \copyright~2025 IEEE. Personal use of this material is permitted. Permission from IEEE must be obtained for all other uses, in any current or future media, including reprinting/republishing this material for advertising or promotional purposes, creating new collective works, for resale or redistribution to servers or lists, or reuse of any copyrighted component of this work in other works.
\end{minipage}}

\title{Interpretable Air Pollution Forecasting by Physics-Guided Spatiotemporal Decoupling
}

\author{
\IEEEauthorblockN{Zhiguo Zhang}
\IEEEauthorblockA{
\textit{KTH Royal Institute of Technology}\\
Stockholm, Sweden \\
zhiguo@kth.se 
}
\and
\IEEEauthorblockN{Xiaoliang Ma\IEEEauthorrefmark{1}}
\IEEEauthorblockA{
\textit{KTH Royal Institute of Technology}\\
Stockholm, Sweden \\
liang@kth.se
}
\and
\IEEEauthorblockN{Daniel Schlesinger}
\IEEEauthorblockA{
\textit{SLB-analys} \\
Stockholm, Sweden \\
daniel.schlesinger@slb.nu
}
}

\AddToShipoutPictureBG*{
  \AtPageUpperLeft{
    \setlength\unitlength{1in}
    \hspace*{\dimexpr0.5\paperwidth\relax}
    \makebox(0,-0.75)[c]{\textbf{Accepted to 2025 IEEE International Conference on Big Data}}
  }
}

\maketitle

\begin{abstract}
Accurate and interpretable air pollution forecasting is crucial for public health, but most models face a trade-off between performance and interpretability. This study proposes a physics-guided, interpretable-by-design spatiotemporal learning framework. 
The model decomposes the spatiotemporal behavior of air pollutant concentrations into two transparent, additive modules. The first is a physics-guided transport kernel with directed weights conditioned on wind and geography (advection). The second is an explainable attention mechanism that learns local responses and attributes future concentrations to specific historical lags and exogenous drivers.
Evaluated on a comprehensive dataset from the Stockholm region, our model consistently outperforms state-of-the-art baselines across multiple forecasting horizons. Our model's integration of high predictive performance and spatiotemporal interpretability provides a more reliable foundation for operational air-quality management in real-world applications.

\end{abstract}


\section{Introduction}
Accurate and reliable urban air pollution forecasts are vital for public health and environmental management. They enable municipal authorities to issue timely health warnings and implement evidence-based traffic management \cite{Crotti2025AURORAE}. 
The evolution of pollutant concentrations across citywide monitoring networks can be described by the advection-diffusion-reaction equation\cite{seinfeld2016atmospheric}. Two processes dominate: i)\textbf{ Spatial transport}, governed mainly by advection along the wind and by near-surface mixing, and ii)\textbf{ Local processes}, arising from site-specific emissions and chemical transformation. Existing modeling paradigms struggle to effectively represent and disentangle these processes, leading to a persistent trade-off between predictive accuracy and model interpretability \cite{zhang2025interpretable, zhu2023machine}.

Prevailing approaches include numerical and data-driven models. Numerical models provide physical fidelity but are constrained by heavy computation and high-resolution inventories, limiting real-time applications \cite{zhang2024improving}. Deep learning models scale well and achieve strong accuracy \cite{zhang2024mgatt, liang2023airformer,zhang2023meta}, yet they compress physical structures into opaque representations with limited diagnostic value. While hybrid physical-neural network designs show promise for improving accuracy \cite{hettige2024airphynet,li2023improving}, they often couple training to costly numerical simulators and their attributions remain opaque.

\begin{figure*}[!t] 
\centering 
\includegraphics[width=1\textwidth]{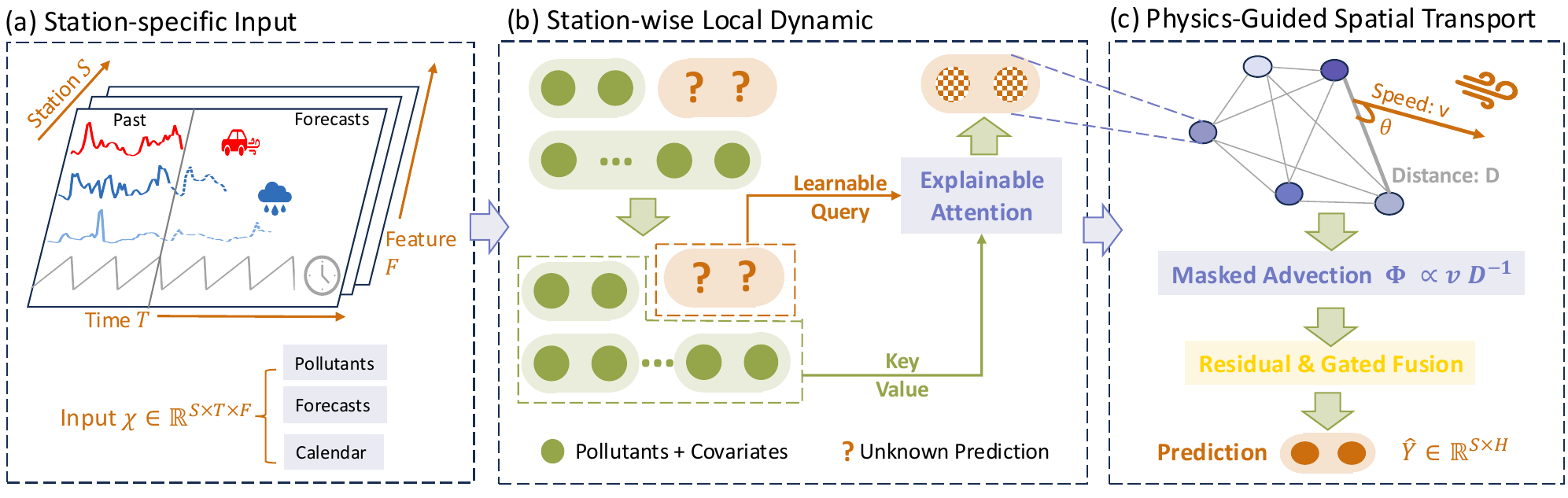} 
\caption{An overview of the proposed model.} 
\label{fig:architecture} 
\end{figure*}

We propose a physics-guided spatiotemporal learning framework that is interpretable-by-design. Our model aligns with physical principles by decoupling the pollutant forecasting into two additive modules with transparent attribution, illustrated in Fig~\ref{fig:architecture} (b) and (c). 
The first module captures station-wise temporal dependencies using a variant of interpretable attention mechanism. A learnable query attends over historical pollutant and auxiliary-variable sequences, attributing the forecast to specific lags and exogenous variables \cite{zhang2025interpretable}. 
The second module learns a physics-guided kernel for spatial transport. Its directed weights are dynamically conditioned on meteorological fields and geographic relationships, forming a learnable spatial operator for advection. This separation provides clear physical meaning to each component and enhances both predictive accuracy and interpretability.

The main contributions are summarized as follows:

\begin{itemize}
  \item An interpretable spatiotemporal framework that separates spatial transport from local dynamics and aligns with physical principles;
  \item A physics-guided, time-varying kernel capturing cross-station interactions underlying the advection process;
  \item A variant of interpretable attention mechanism providing station-level attributions that map each input variable to its contribution at each forecast step;
  \item Extensive experiments on a city-scale case study demonstrate superior accuracy over state-of-the-art baselines and yield spatiotemporal interpretations validated through a case analysis.
\end{itemize}

\section{Related Work}
\subsection{Spatiotemporal Forecasting} 
Numerical air quality models, spanning from regional Chemical Transport Models (CTMs) \cite{cams2017reg} to local street-level models \cite{ottosen2015analysis}, provide physical grounding. However, they are computationally intensive and highly sensitive to the accuracy of emission inventories and meteorological inputs, limiting their operational agility in complex urban environments \cite{zhang2023meta}.
\IEEEpubidadjcol
Data-driven methods, including Recurrent Neural Networks (RNNs), Convolutional Neural Networks (CNNs), Graph Neural Networks (GNNs), and Transformers, excel at modeling complex nonlinear dependencies \cite{ wu2020connecting, liang2023airformer, vaswani2017attention}. A persistent challenge, however, is interpretability. Many GNNs employ static, distance-based graphs, which are a poor surrogate for wind-driven atmospheric transport \cite{zhang2024mgatt}. More advanced dynamic graph models often derive connectivity from statistical correlations, resulting in opaque representations that are disconnected from underlying physical processes like advection.

\subsection{Physics-Guided Models}
To build more trustworthy models, researchers are integrating physical knowledge into neural networks. One approach, Physics-Informed Neural Networks (PINNs), uses governing PDEs as soft constraints \cite{raissi2019physics}. This guides the model toward physically plausible solutions but does not render the network architecture inherently interpretable. Another direction involves hybrid models that use neural networks to parameterize components of a physical process, often formulated with Ordinary Differential Equations (ODEs). For instance, AirPhyNet uses a GNN-based differential equation network to model the physical transport of air particles, specifically diffusion and advection\cite{hettige2024airphynet}. While these methods improve physical consistency, their reliance on iterative ODE solvers can introduce significant computational overhead during training. Furthermore, their intricate coupling of neural and physical components can still obscure clear attribution.

\section{Preliminaries}
\subsection{Problem Formulation}
We forecast PM$_{10}$ at a network of $S$ stations over a future horizon of $H$ hours, conditioned on a look-back window of $L$ hours. All series are sampled hourly.

Features are indexed by $i\in\{1,\dots,F\}$ and stations by $s\in\{1,\dots,S\}$. At timestep $t$, we define $\mathbf{z}^{i,s}_{t-L+1:t+H}\in\mathbb{R}^{L+H}$ as the subsequence of feature $i$ at station $s$ from $t\!-\!L\!+\!1$ to $t\!+\!H$. Stack the $F$ features to form the station-level input
\[
\mathbf{x}^s=\big[\,\mathbf{z}^{1,s}\ \mathbf{z}^{2,s}\ \cdots\ \mathbf{z}^{F,s}\,\big]\in\mathbb{R}^{(L+H)\times F},
\]
and stack stations to obtain the multi-station input
\[
\mathcal{X}=\big[\,\mathbf{x}^1,\mathbf{x}^2,\dots,\mathbf{x}^S\,\big]\in\mathbb{R}^{S\times(L+H)\times F}.
\]
This input $\mathcal{X}$ contains historical observations and exogenous covariates with varying future availability $m_i \le H$, such as meteorological forecasts and calendar features \cite{zhang2025interpretable}.

The target is $\mathbf{Y}\in\mathbb{R}^{S\times H}$, where $Y_{s,h}$ denotes PM$_{10}$ concentration at station $s$ and lead time $h\in\{1,\dots,H\}$. The learning objective is to train a model $f_{\Theta}$ with parameters $\Theta$ that maps the heterogeneous inputs to multi-horizon forecasts:
\begin{equation}
\widehat{\mathbf{Y}} = f_{\Theta}\!\big(\mathcal{X}\big)\in\mathbb{R}^{S\times H}.
\end{equation}

\subsection{Governing Equations and Model Formulation}
Let $c(\mathbf{x},t)$ be the near-surface PM$_{10}$ concentration at location $\mathbf{x}$ and time $t$. Its evolution is described by the advection-diffusion-reaction equation \cite{seinfeld2016atmospheric}:
\begin{equation}
\frac{\partial c}{\partial t} \;+\; \mathbf{u}(\mathbf{x},t)\cdot\nabla c
\;=\; \nabla\!\cdot\!\big(\mathbf{K}(\mathbf{x},t)\nabla c\big) \;+\; \mathcal{R}\!\big(c,\chi(\mathbf{x},t)\big) \;+\; S(\mathbf{x},t),
\label{eq:ADR}
\end{equation}
where $\mathbf{u}$ is the wind field, $\mathbf{K}$ is the turbulent diffusivity tensor, $\mathcal{R}$ represents chemical reactions and deposition under environmental state $\chi$, and $S$ denotes emissions and re-suspension.

For urban PM$_{10}$ forecasting, non-local transport between monitoring sites is primarily governed by advection driven by the wind field $\mathbf{u}$ \cite{stull2012introduction}. In contrast, processes such as turbulent mixing ($\mathbf{K}$), chemical reactions ($\mathcal{R}$), and emissions ($S$) operate at a more local scale \cite{zhang2001size, seinfeld2016atmospheric}. This physical scale separation motivates decomposing the PM$_{10}$ forecast $Y_{s,h}$ at a station $s$ and lead time $h$ into additive advection and local components:
\begin{equation}
Y_{s,h}
\;=\;
\underbrace{\sum_{s'} W^{\mathrm{adv}}_{h}(s,s')\,\phi_{s',h}}_{\text{advection-dominated transport}}
\;+\;
\underbrace{L_{s,h}(\mathcal{X})}_{\text{local dynamics}}
\;+\;\varepsilon_{s,h}.
\label{eq:local_adv}
\end{equation}

Here, $W^{\mathrm{adv}}_{h}(s,s')$ is a directed kernel representing wind-driven transport from a source station $s'$ to a target station $s$; $\phi_{s',h}$ is the transported signal from the source station; $L_{s,h}(\mathcal{X})$ is a local dynamics term capturing site-specific processes informed by the covariates $\mathcal{X}$; and $\varepsilon_{s,h}$ is the residual error.

\section{Methodology}
\begin{figure}[!t] \centering 
\includegraphics[width=0.45\textwidth]{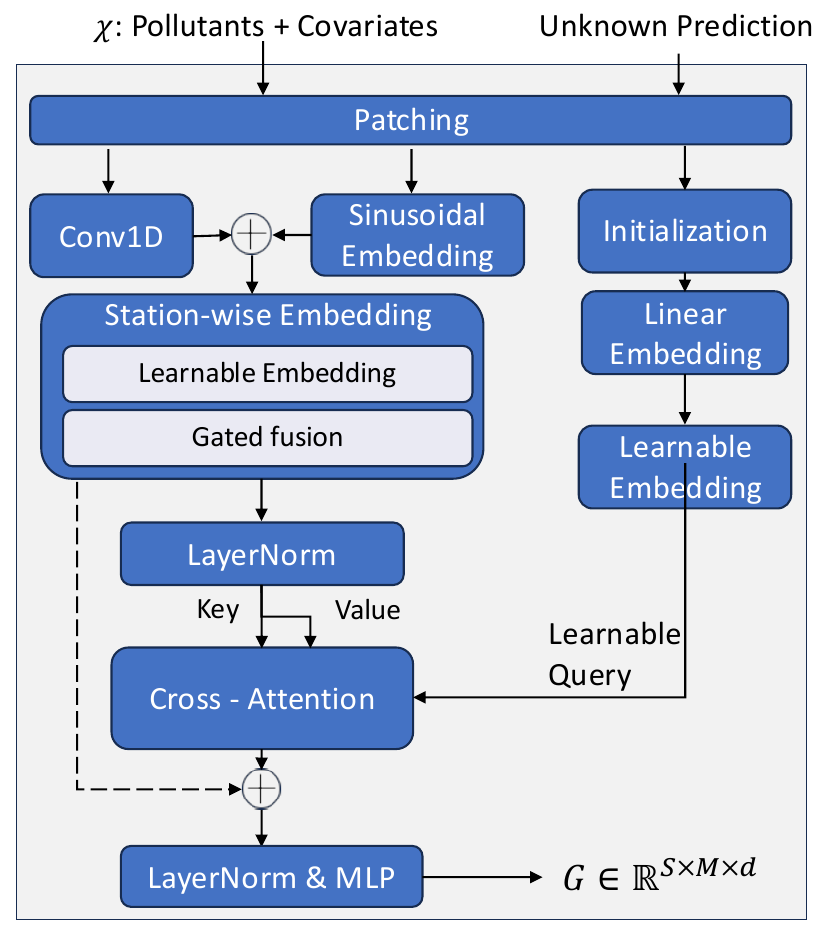} 
\caption{The embedding and attention mechanism for modeling local dynamics.} 
\label{fig:plot1} 
\end{figure}

Our framework proposes an interpretable-by-design architecture that structurally decouples the spatiotemporal forecasting into two distinct components. The first is a station-wise attention-based encoder for modeling local dynamics, illustrated in Fig~\ref{fig:plot1}. The second is a physics-guided module for modeling external transport, shown in Fig~\ref{fig:plot2}.

\subsection{Embedding of Time-Feature Tokens}
For each of the $F$ input features at a given station, we first tokenize the time series by partitioning it into non-overlapping patches of length $P$. Each patch constitutes a single semantic unit, termed a \emph{time-feature token}. This tokenization is applied to all available inputs, yielding $N_{\mathrm{all}}$ tokens per station.

We project the $N_{\mathrm{all}}$ time-feature tokens into a $d$-dimensional space and augment them with 1D convolutional and sinusoidal positional embeddings ($\mathbf{P}^{\mathrm{conv1D}}$, $\mathbf{P}^{\mathrm{feat}}$) to form $\mathbf{H}^{(s)}$ \cite{liang2025selection,zhang2023crossformer}.
\begin{equation}
\mathbf{H}^{(s)} = \Phi\!\big(\mathrm{tokenize}(\mathcal{X}_{s})\big) + \mathbf{P}^{\mathrm{conv1D}} + \mathbf{P}^{\mathrm{feat}}.
\label{eq:H_s}
\end{equation}
Subsequently, we integrate a learnable station embedding $\mathbf{E}$ via a station-wise affine gating mechanism \cite{chi2023difforecast} to produce the final input representation $\widetilde{\mathbf{H}} \in \mathbb{R}^{S\times N_{\mathrm{all}}\times d}$.
\begin{equation}
\widetilde{\mathbf{H}} = \mathbf{\Gamma}\odot\big(\mathbf{H}+\mathbf{E}\big) + \mathbf{B},
\label{eq:H_tilde}
\end{equation}
where $\mathbf{\Gamma}, \mathbf{B}\in\mathbb{R}^{S\times d}$ are learnable gating parameters. The resulting tensor $\widetilde{\mathbf{H}}$ serves as the source for keys and values in the subsequent attention module.

\subsection{Station-wise Local Dynamics Encoder}
The station-wise encoder models the temporal dynamics unique to each monitoring station. A key feature of this module is its ability to yield interpretable, station-wise attributions that link each forecasted value to specific input features over time.

This is achieved through a variant of X2-attention mechanism\cite{zhang2025interpretable}. The keys ($\mathbf{K}$) and values ($\mathbf{V}$) are derived from $\widetilde{\mathbf{H}}$ via linear projections.
\begin{equation}
\begin{aligned}
\mathbf{K} &= \widetilde{\mathbf{H}}\,\mathbf{W}_{K}
\in \mathbb{R}^{S\times N_{\mathrm{all}}\times d},\\
\mathbf{V} &= \widetilde{\mathbf{H}}\,\mathbf{W}_{V}
\in \mathbb{R}^{S\times N_{\mathrm{all}}\times d}.
\end{aligned}
\label{eq:KV}
\end{equation}
In contrast to standard self-attention, the queries ($\mathbf{Q}$) are not derived from the input data but function as learnable representatives for the forecast targets. Let $M_{\mathrm{pred}}$ be the number of future PM$_{10}$ patches to forecast. The initial query embeddings, $\mathbf{Q}^{\mathrm{emb}}$, are constructed from learnable temporal and feature-specific positional encodings.

\begin{equation}
\mathbf{Q}^{\mathrm{emb}} = \mathbf{P}^{\mathrm{time}}_{\mathrm{pred}} + \mathbf{P}^{\mathrm{feat}}_{\mathrm{pred}}
\in \mathbb{R}^{S\times M_{\mathrm{pred}}\times d}.
\label{eq:Q_emb}
\end{equation}
\begin{equation}
\mathbf{Q} = \mathbf{Q}^{\mathrm{emb}}\,\mathbf{W}_{Q}\in\mathbb{R}^{S\times M_{\mathrm{pred}}\times d}.
\label{eq:Q}
\end{equation}

The attention scores $\mathbf{A}$ and the resulting context vectors $\mathbf{G}$ are computed as:

\begin{equation}
\begin{aligned}
\mathbf{A} &= \mathrm{softmax}\!\left(\frac{\mathbf{Q}\mathbf{K}^{\top}}{\sqrt{d_{k}}}\right)
\in \mathbb{R}^{S \times M_{\mathrm{pred}} \times N_{\mathrm{all}}},\\
\mathbf{G} &= \mathbf{A}\,\mathbf{V}
\in \mathbb{R}^{S \times M_{\mathrm{pred}} \times d}.
\end{aligned}
\label{eq:Attention}
\end{equation}
The attention parameters ($\mathbf{W}_{Q}, \mathbf{W}_{K}, \mathbf{W}_{V}$) are shared across all $S$ stations, enabling the module to learn a generalized model of local temporal dynamics. The output $\mathbf{G}$ corresponds to the local dynamics term, $L_{s,h}(\mathcal{X})$, in Eq.~\ref{eq:local_adv}. This tensor encapsulates the forecast component attributable to site-specific processes and is subsequently fused with the output of the physics-guided transport module.

\begin{figure}[!t] \centering 
\includegraphics[width=0.5\textwidth]{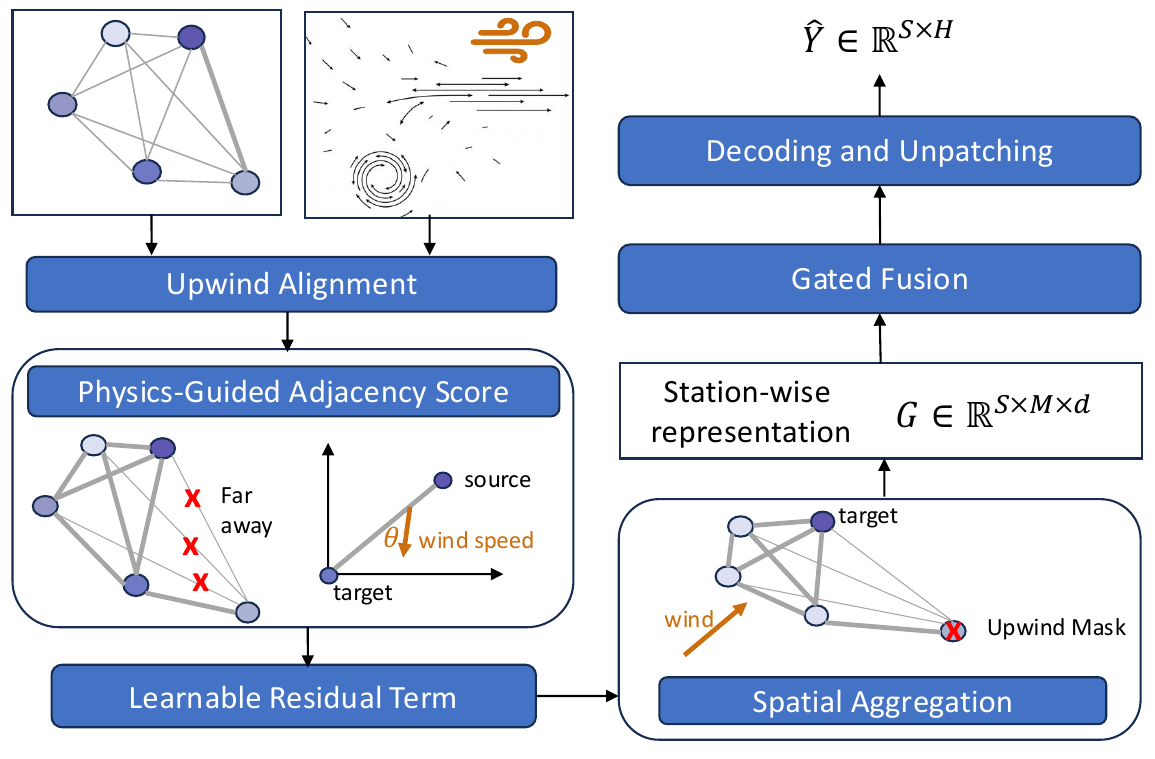} 
\caption{The physics-guided module for cross-station transport.} 
\label{fig:plot2} 
\end{figure}

\subsection{Physics-Guided Cross-Station Transport}
The spatial propagation of pollutants is modeled by a physics-guided transport module. This component refines the local dynamics representations, $\mathbf{G}$, by aggregating information from neighboring stations based on meteorological conditions and geographical relationships, as depicted in Fig~\ref{fig:plot2}.

\paragraph{Local Geometry and Upwind Alignment}
We first compute the pairwise station distance $\mathbf{D}$ and unit bearing vectors $\widehat{\mathbf{R}}$. The wind alignment score $(\mathbf{A}^{\mathrm{wind}}_m)$ between each station pair and the forecasted wind vector $\widehat{\mathbf{u}}_{m}$ is then calculated in Eq.~\ref{eq:wind_align}.
\begin{equation}
(\mathbf{A}^{\mathrm{wind}}_m)_{s,s_0}
=\widehat{\mathbf{u}}_m^{\top}\,\widehat{\mathbf{R}}_{s,s_0}\in[-1,1].
\label{eq:wind_align}
\end{equation}
A positive value indicates that the target station $s$ is downwind of the source station $s_0$, making $s_0$ a potential source of advected pollutants.

\paragraph{Physics-Guided Adjacency Score}
We formulate an unnormalized physics score, $\boldsymbol{\Phi}_m \in \mathbb{R}^{S\times S}$, that encourages upwind influence and penalizes distance. For each forecast patch $m$, this score is:
\begin{equation}
\boldsymbol{\Phi}_m
=\beta_{\text{speed}}\, v_m \cdot \alpha_{\mathrm{dir}}\,[\,\mathbf{A}^{\mathrm{wind}}_m-\varepsilon\,]_+
\;-\;\alpha_{\mathrm{dist}}\Big(\tfrac{\mathbf{D}}{\sigma_d}\Big)^{\!2},
\end{equation}
where $[\cdot]_+ = \max(0, \cdot)$ is the rectifier function and $v_m$ is the wind speed. The coefficients $\alpha_{\text{dir}}$, $\alpha_{\text{dist}}$, $\beta_{\text{speed}}$, the distance scale $\sigma_d$, and the alignment margin $\varepsilon$ are learnable parameters that allow the model to calibrate the physical formulation.

To capture static spatial dependencies not governed by meteorology, such as geographical barriers or fixed emission corridors, we add a learnable residual adjacency matrix $\mathbf{U} \in \mathbb{R}^{S \times S}$. The final adjacency score is:
\begin{equation}
\mathbf{Z}_m=\boldsymbol{\Phi}_m+\mathbf{U}.
\end{equation}

\paragraph{Upwind Spatial Aggregation}
To strictly enforce the principle of advection, we ensure that pollutant transport originates only from upwind sources. This is achieved using a dynamic binary mask $\mathbf{M}_m \in \{0,1\}^{S \times S}$. An entry in the mask is set to 1 if a source station $s_0$ is upwind of a target station $s$, defined by the condition $(\mathbf{A}^{\mathrm{wind}}_m)_{s,s_0} \ge \varepsilon$. All non-upwind pairs and self-loops ($s=s_0$) are masked out.

For each forecast horizon $m$, we compute dynamic mixing weights by applying this mask to the adjacency scores and normalizing the result row-wise:
\begin{equation}
\mathbf{W}^{\mathrm{sp}}_m = \mathrm{Softmax}_{\text{row}}\!\big(\,\mathbf{Z}_m\odot \mathbf{M}_m\,\big) \in \mathbb{R}^{S\times S}.
\end{equation}
The resulting weight matrix $\mathbf{W}^{\mathrm{sp}}_m$ defines the influence of each station on every other station for that specific time step, satisfying $\sum_{s'=1}^{S}(\mathbf{W}^{\mathrm{sp}}_m)_{s,s'}=1$ for each target station $s$, with $(\mathbf{W}^{\mathrm{sp}}_m)_{s_0,s_0}=0$.

Finally, the cross-station transport representation, $\mathbf{C}_{\mathrm{nb}}$, is obtained by aggregating the local dynamics representations $\mathbf{G}$ using these spatial weights. For each future patch $m$, this is computed as:
\begin{equation}
\mathbf{C}_{\mathrm{nb}}[:,m,:] = \mathbf{W}^{\mathrm{sp}}_{m}\;\mathbf{G}[:,m,:] \in \mathbb{R}^{S\times d}.
\end{equation}
Stacking these representations across all future patches yields the final transported context tensor $\mathbf{C}_{\mathrm{nb}}\in\mathbb{R}^{S\times M_{\mathrm{pred}}\times d}$.

\subsection{Fusion and Decoding}
The final forecast is generated by fusing the local and transported representations, which are then mapped back to the target temporal resolution through a decoding head. Specifically, the local representation $\mathbf{G}$ and the transported representation $\mathbf{C}_{\mathrm{nb}}$ are fused via a station-specific gate $\boldsymbol{\gamma}$.
\begin{equation}
\mathbf{C} = \mathbf{G} + \boldsymbol{\gamma} \odot \mathbf{C}_{\mathrm{nb}} \in \mathbb{R}^{S\times M_{\mathrm{pred}}\times d}.
\label{eq:C_fused}
\end{equation}
A non-linear decoding head then maps the fused representation $\mathbf{C}$ back to predictive patches.
\begin{equation}
\widehat{\mathbf{Y}}_{\mathrm{patch}}
= \sigma(\mathbf{C}\,\mathbf{W}_{\text{dec}} + \mathbf{b}_{\text{dec}})
\in \mathbb{R}^{S\times M_{\mathrm{pred}}\times P}.
\label{eq:Y_patch}
\end{equation}
These patches are concatenated and reshaped to form the final multi-horizon forecast tensor $\widehat{\mathbf{Y}}$.
\begin{equation}
\widehat{\mathbf{Y}} = \mathrm{Reshape}(\widehat{\mathbf{Y}}_{\mathrm{patch}}) \in \mathbb{R}^{S\times H}.
\label{eq:Y_hat}
\end{equation}

\subsection{Learning Objective}
The model parameters are optimized end-to-end by minimizing a composite loss function. The primary objective is the Mean Squared Error (MSE). This is augmented with a mild $\ell_2$ regularization on the learnable margin parameter $\varepsilon$ to stabilize its value during training:
\begin{equation}
\mathcal{L} = \frac{1}{SH}\sum_{s=1}^{S}\sum_{h=1}^{H}\big(\hat{y}_{s,h}-y_{s,h}\big)^2 + \lambda_{\varepsilon}\,\varepsilon^{2}.
\end{equation}

\subsection{Spatiotemporal Interpretability}
A central contribution is that our model provides two complementary and physically meaningful forms of attribution.

First, \textbf{Temporal-Feature Interpretability} is provided by the attention maps $\mathbf{A}$ and quantify the influence of each historical time-feature token on each future prediction, revealing which past observations and covariates are most salient for the forecast at a specific station.

Second, \textbf{Spatial Interpretability} is achieved through the physics-guided transport module. The dynamic spatial weights, $\mathbf{W}^{\mathrm{sp}}_m$, offer a clear, horizon-specific attribution of how much influence each upwind station contributes to a target station's forecast. This is complemented by the station-specific gates, $\boldsymbol{\gamma}$, which quantify the overall importance of advective transport relative to local dynamics for each station.

\section{Experiments}
\subsection{Dataset}
Our study utilizes hourly air quality data at nine stations from the Air Quality Forecast System in Stockholm, Sweden\cite{engardt2021luften}. The dataset spans from July 2020 to October 2024 and is chronologically partitioned into training (70\%), validation (20\%), and testing (10\%) sets.

To improve predictive accuracy, we incorporate a range of auxiliary variables, including meteorological forecasts, deterministic air quality forecasts from a multi-scale coupled modeling system, and temporal encodings for cyclical patterns. Further details on these auxiliary features and their integration are provided in \cite{zhang2024improving}

\subsection{Baselines}
We evaluate our model against a diverse set of state-of-the-art baselines. These include AirPhyNet \cite{hettige2024airphynet}, a hybrid physics-guided model, and Airformer \cite{liang2023airformer}, a Transformer architecture tailored for air quality forecasting. We also compare against general-purpose multivariate forecasting models: Crossformer \cite{zhang2023crossformer}, which employs two-stage attention, and iTransformer \cite{liu2023itransformer}, which inverts the attention mechanism to capture variate-wise correlations. The foundational encoder-only Transformer \cite{vaswani2017attention} serves as a fundamental benchmark. While these models are effective predictors, none are architecturally designed to provide the decoupled spatiotemporal interpretability that our framework offers.

\begin{table*}[!t]
\caption{Model performance for PM\textsubscript{10} at 9 monitoring stations. The best results are highlighted in bold, second-best results are underlined, and Avg denotes the mean across all forecast horizons (24, 48, 72 hours).}
\label{tab:long_term_forecasting_results}
\centering
\begin{tabular}{c|c|cc|cc|cc|cc|cc|cc}
    \toprule
    \multicolumn{2}{c|}{\textbf{Models}} &
      \multicolumn{2}{c|}{\textbf{Ours}} &
      \multicolumn{2}{c|}{Airformer} &
      \multicolumn{2}{c|}{AirPhyNet} &
      \multicolumn{2}{c|}{Crossformer} &
      \multicolumn{2}{c|}{iTransformer} &
      \multicolumn{2}{c}{Transformer} \\

    \multicolumn{2}{c|}{} &
      \multicolumn{2}{c|}{(this work)} &
      \multicolumn{2}{c|}{(2023)} &
      \multicolumn{2}{c|}{(2024)} &
      \multicolumn{2}{c|}{(2023)} &
      \multicolumn{2}{c|}{(2023)} &
      \multicolumn{2}{c}{(2017)} \\
      
    \cmidrule(lr){3-4}\cmidrule(lr){5-6}\cmidrule(lr){7-8}
    \cmidrule(lr){9-10}\cmidrule(lr){11-12}\cmidrule(lr){13-14}
    \multicolumn{2}{c|}{\textbf{Metric}} & MAE & MSE & MAE & MSE & MAE & MSE & MAE & MSE & MAE & MSE & MAE & MSE \\
    \midrule
    \multirow{4}{*}{Horizon} 
        & 24  & \textbf{4.71} & \textbf{76.97} & 5.13 & 92.23 & 5.15 & \underline{92.21}  & 5.62 & 96.59 & \underline{5.03} & 98.58 & 5.95 & 94.28 \\
        & 48  & \textbf{4.95} & \textbf{85.13} & 5.64 & 102.72 & \underline{5.63} & 99.47  & 5.66 & \underline{95.93} & 5.72 & 105.56 & 6.00 & 96.55  \\
        & 72  & \textbf{5.26} & \textbf{90.44} & 5.68 & 98.82 & 5.86 & 104.07 & 6.11 & 104.65 & 5.90 & 113.75 & \underline{5.64} & \underline{95.41} \\
         & AVG & \textbf{4.97} & \textbf{84.18} & \underline{5.48} & 97.92 & 5.55 & 98.58  & 5.80 & 99.06 & 5.55 & 105.96 & 5.86 & \underline{95.41}\\
    \bottomrule
  \end{tabular}
\end{table*}

\begin{figure*}[!t] 
\centering 
\includegraphics[width=0.8\textwidth]{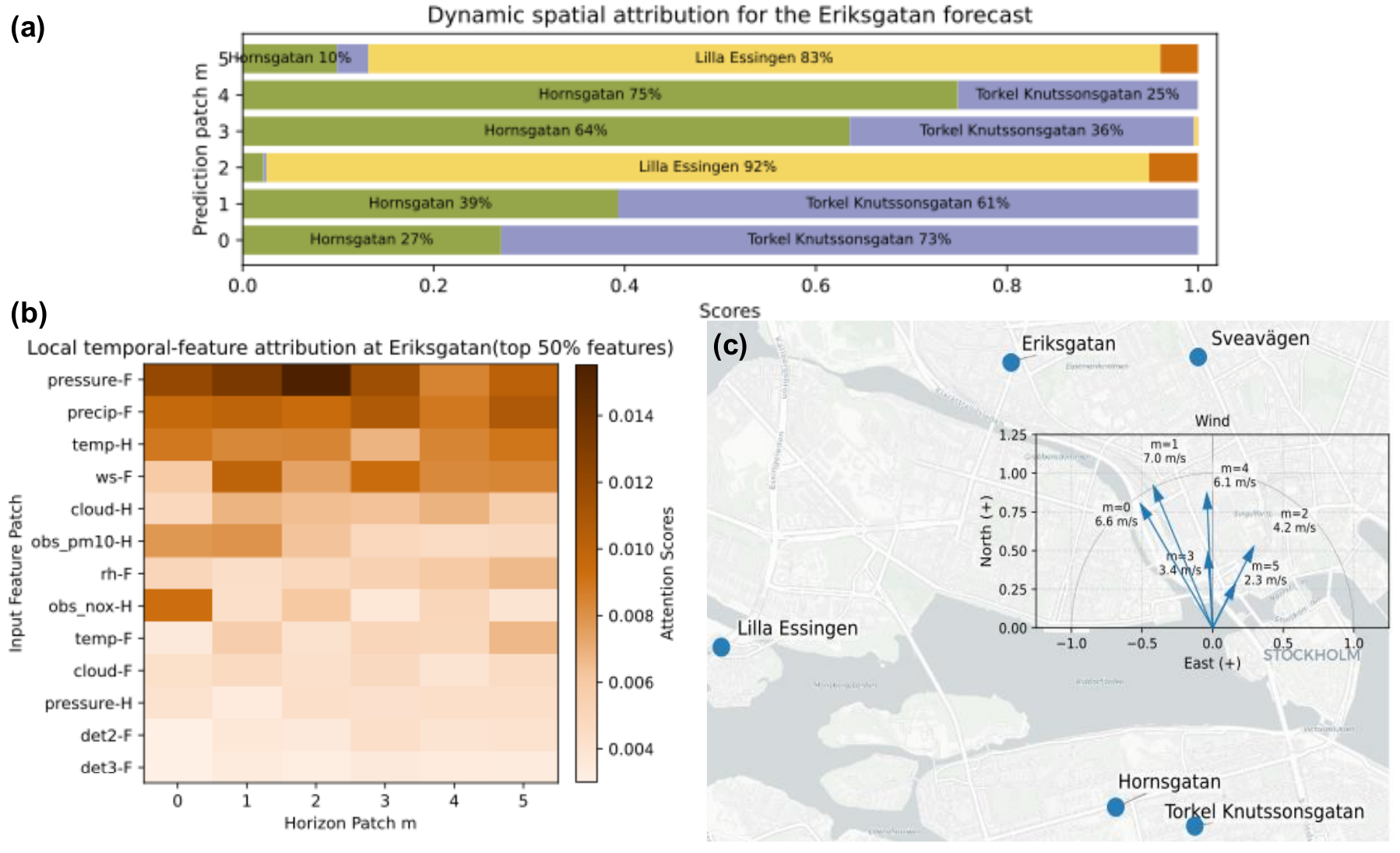} 
\caption{(a) Dynamic spatial attribution for the S:t Eriksgatan forecast. Each bar represents a 12-hour forecast patch, with the percentage contribution from each upwind station, as determined by the learned spatial weights $\mathbf{W}^{\mathrm{sp}}_m$; (b) Local temporal-feature attribution for a 72-hour forecast at S:t Eriksgatan. The color intensity indicating the influence of each input feature patch on each future prediction patch; (c) The distribution of wind direction and speed, serving as meteorological ground truth to validate the learned spatial attributions in (a).}
\label{fig:local_attribution} 
\end{figure*} 

\subsection{Implementation Details}

All models were implemented in PyTorch and trained on a single NVIDIA A40 GPU. We minimized MSE loss using the Adam optimizer and employed an early stopping strategy. We evaluated models across forecast horizons $H \in \{24, 48, 72\}$ hours, setting the look-back window to $L = H + 24$ hours and using a fixed patch length of 12 hours. Hyperparameters, such as the embedding dimension $d$ and the number of attention heads $n_{\text{heads}}$, were optimized using the Optuna framework. To ensure a fair comparison, baseline Transformer models were configured with a single encoder and decoder layer. Model performance is evaluated using Mean Absolute Error (MAE) and MSE.

\subsection{Result analysis}
\noindent\textbf{Model Performance.}
Table~\ref{tab:long_term_forecasting_results} presents the comparative performance of our model against the baselines. Our proposed framework achieves the lowest MAE and MSE across all 24, 48, and 72-hour forecast horizons.

On average, our model reduces MAE by 9.3\% and MSE by 14.0\% relative to Airformer, the strongest domain-specific baseline. This consistent improvement highlights the efficacy of our physics-guided, decoupled architecture. While domain-adapted models like AirPhyNet generally outperform general-purpose Transformer variants, vanilla Transformer shows surprisingly competitive performance at the 72-hour horizon, surpassing other baselines. Nevertheless, our model still outperforms it by an average MSE margin of 11.8\%.

\noindent\textbf{Spatial-temporal Interpretability.}
A key strength of our framework is its ability to provide fine-grained spatiotemporal attributions. We demonstrate this with a case study of a 72-hour forecast for the centrally located S:t Eriksgatan station.

As shown in Figure~\ref{fig:local_attribution} (b), local dynamics, interpreted via attention scores, reveal past PM\textsubscript{10}, pressure, and precipitation as most influential. The scores decay over the forecast horizon, correctly reflecting diminishing influence of older information.

Advective transport, interpreted via dynamic weights $\mathbf{W}^{\mathrm{sp}}_m$, is shown in Figure~\ref{fig:local_attribution} (a). The results are validated against meteorological data from the same period, shown in Figure~\ref{fig:local_attribution} (c). For instance, during south-easterly winds (patches 0, 1, 3, 4), the model correctly attributes influence to the upwind Hornsgatan and Torkel Knutssonsgatan stations. When the wind shifts to south-westerly (patches 2, 5), the model dynamically assigns dominant contributions (92\% and 83\%) to the new upwind source, Lilla Essingen. This meteorology-consistent attribution confirms the model learned a physically meaningful transport representation.

\section{Conclusion}
In this paper, we propose a physics-guided spatiotemporal learning framework for air-quality forecasting that prioritizes interpretability without sacrificing accuracy. By structurally decoupling the evolution of air pollutant concentrations into spatial transport and local responses, the model aligns with core physical principles. A dynamic, wind-conditioned transport kernel explicitly models the advection processes across monitoring sites, while a station-wise interpretable attention encoder provides fine-grained attributions to historical lags and exogenous variables. 

Extensive experiments on real-world data show that this design consistently outperforms state-of-the-art baselines across multiple forecasting horizons, and yields actionable spatiotemporal insights. Overall, coupling an explainable neural architecture with a learnable physics prior provides a practical path to accurate and trustworthy city-scale air-quality forecasts.

\section*{Acknowledgment}
The authors thank the Richterska Stiftelsen (2024-00992) and the KTH Centre for Digital Futures (iHorse+ 2024) for supporting this work. The authors also acknowledge Chalmers e-Commons at Chalmers University of Technology, and the National Academic Infrastructure for Supercomputing in Sweden (NAISS) (funded by Swedish Research Council grant 2022-06725) for providing the computational resources.

\bibliographystyle{IEEEtran}   
\bibliography{refs}

@article{zhang2025interpretable,
  title   = {Interpretable Long-Horizon Air Pollutants Forecasting using A Transformer-based Framework},
  author  = {Zhang, Zhiguo and Ma, Xiaoliang and Engardt, Magnuz and Schlesinger, Daniel and Johansson, Christer},
  year    = {2025},
  journal = {SSRN Electronic Journal},
  doi     = {10.2139/ssrn.5658909},
  url     = {https://ssrn.com/abstract=5658909}
}

@inproceedings{wu2020connecting,
  title={Connecting the dots: Multivariate time series forecasting with graph neural networks},
  author={Wu, Zonghan and Pan, Shirui and Long, Guodong and Jiang, Jing and Chang, Xiaojun and Zhang, Chengqi},
  booktitle={Proceedings of the 26th ACM SIGKDD international conference on knowledge discovery \& data mining},
  pages={753--763},
  year={2020}
}

@article{vaswani2017attention,
  title={Attention is all you need},
  author={Vaswani, Ashish and Shazeer, Noam and Parmar, Niki and Uszkoreit, Jakob and Jones, Llion and Gomez, Aidan N and Kaiser, {\L}ukasz and Polosukhin, Illia},
  journal={Advances in neural information processing systems},
  volume={30},
  year={2017}
}

@article{raissi2019physics,
  title={Physics-informed neural networks: A deep learning framework for solving forward and inverse problems involving nonlinear partial differential equations},
  author={Raissi, Maziar and Perdikaris, Paris and Karniadakis, George E},
  journal={Journal of Computational physics},
  volume={378},
  pages={686--707},
  year={2019},
  publisher={Elsevier}
}

@misc{cams2017reg,
  author={Copernicus Atmosphere Monitoring Service},
  title={Meteo-France: Regional Production,Description of the operational models and of the ENSEMBLE system},
  howpublished={\url{https://atmosphere.copernicus.eu/sites/default/files/2018-02/CAMS50_factsheet_201610_v2.pdf}},
  note         = {Last accessed: 15 June 2025},
  year={2017},
}

@inproceedings{liang2023airformer,
  title={Airformer: Predicting nationwide air quality in china with transformers},
  author={Liang, Yuxuan and Xia, Yutong and Ke, Songyu and Wang, Yiwei and Wen, Qingsong and Zhang, Junbo and Zheng, Yu and Zimmermann, Roger},
  booktitle={Proceedings of the AAAI conference on artificial intelligence},
  volume={37},
  number={12},
  pages={14329--14337},
  year={2023}
}

@article{hettige2024airphynet,
  title={Airphynet: Harnessing physics-guided neural networks for air quality prediction},
  author={Hettige, Kethmi Hirushini and Ji, Jiahao and Xiang, Shili and Long, Cheng and Cong, Gao and Wang, Jingyuan},
  journal={arXiv preprint arXiv:2402.03784},
  year={2024}
}

@article{liu2023itransformer,
  title={itransformer: Inverted transformers are effective for time series forecasting},
  author={Liu, Yong and Hu, Tengge and Zhang, Haoran and Wu, Haixu and Wang, Shiyu and Ma, Lintao and Long, Mingsheng},
  journal={arXiv preprint arXiv:2310.06625},
  year={2023}
}

@inproceedings{zhang2023crossformer,
  title={Crossformer: Transformer utilizing cross-dimension dependency for multivariate time series forecasting},
  author={Zhang, Yunhao and Yan, Junchi},
  booktitle={The eleventh international conference on learning representations},
  year={2023}
}

@book{seinfeld2016atmospheric,
  title={Atmospheric chemistry and physics: from air pollution to climate change},
  author={Seinfeld, John H and Pandis, Spyros N},
  year={2016},
  publisher={John Wiley \& Sons}
}

@inproceedings{zhang2023meta,
  title={A meta-graph deep learning framework for forecasting air pollutants in Stockholm},
  author={Zhang, Zhiguo and Ma, Xiaoliang and Johansson, Christer and Jin, Junchen and Engardt, Magnuz},
  booktitle={2023 IEEE 9th World Forum on Internet of Things (WF-IoT)},
  pages={01--06},
  year={2023},
  organization={IEEE}
}

@article{zhang2024improving,
  title={Improving 3-day deterministic air pollution forecasts using machine learning algorithms},
  author={Zhang, Zhiguo and Johansson, Christer and Engardt, Magnuz and Stafoggia, Massimo and Ma, Xiaoliang},
  journal={Atmospheric Chemistry and Physics},
  volume={24},
  number={2},
  pages={807--851},
  year={2024},
  publisher={Copernicus GmbH}
}

@article{zhang2024mgatt,
  title={MGAtt-LSTM: A multi-scale spatial correlation prediction model of PM2. 5 concentration based on multi-graph attention},
  author={Zhang, Bo and Chen, Weihong and Li, Mao-Zhen and Guo, Xiaoyang and Zheng, Zhonghua and Yang, Ru},
  journal={Environmental Modelling \& Software},
  volume={179},
  pages={106095},
  year={2024},
  publisher={Elsevier}
}

@article{li2023improving,
  title={Improving air quality assessment using physics-inspired deep graph learning},
  author={Li, Lianfa and Wang, Jinfeng and Franklin, Meredith and Yin, Qian and Wu, Jiajie and Camps-Valls, Gustau and Zhu, Zhiping and Wang, Chengyi and Ge, Yong and Reichstein, Markus},
  journal={npj Climate and Atmospheric Science},
  volume={6},
  number={1},
  pages={152},
  year={2023},
  publisher={Nature Publishing Group UK London}
}

@techreport{engardt2021luften,
  author       = {Engardt, Magnuz and Bergström, Sebastian and Johansson, Christer},
  title        = {Luften du andas -- nu och de kommande dagarna: Utveckling av ett automatiskt prognossystem för luftföroreningar och pollen},
  institution  = {SLB-analys, Stockholm Stad},
  year         = {2021},
  number       = {SLB 36:2021},
  pages        = {33},
  note         = {In Swedish}
}

@book{stull2012introduction,
  title={An introduction to boundary layer meteorology},
  author={Stull, Roland B},
  volume={13},
  year={2012},
  publisher={Springer Science \& Business Media}
}

@article{zhang2001size,
  title={A size-segregated particle dry deposition scheme for an atmospheric aerosol module},
  author={Zhang, Leiming and Gong, Sunling and Padro, Jacob and Barrie, Len},
  journal={Atmospheric environment},
  volume={35},
  number={3},
  pages={549--560},
  year={2001},
  publisher={Elsevier}
}

@article{ottosen2015analysis,
  title={Analysis of the impact of inhomogeneous emissions in the Operational Street Pollution Model (OSPM)},
  author={Ottosen, T-B and Kakosimos, KE and Johansson, Christer and Hertel, Ole and Brandt, J{\o}rgen and Skov, Henrik and Berkowicz, Ruwim and Ellermann, Thomas and Jensen, Steen Solvang and Ketzel, Matthias},
  journal={Geoscientific Model Development},
  volume={8},
  number={10},
  pages={3231--3245},
  year={2015},
  publisher={Copernicus GmbH G{\"o}ttingen, Germany}
}

@article{liang2025selection,
  title={Selection of Layers from Self-supervised Learning Models for Predicting Mean-Opinion-Score of Speech},
  author={Liang, Xinyu and Cumlin, Fredrik and Ungureanu, Victor and Reddy, Chandan KA and Schuldt, Christian and Chatterjee, Saikat},
  journal={arXiv preprint arXiv:2508.08962},
  year={2025}
}

@inproceedings{chi2023difforecast,
  title={Difforecast: Image generation based highway traffic forecasting with diffusion model},
  author={Chi, Pengnan and Ma, Xiaoliang},
  booktitle={2023 IEEE International Conference on Big Data (BigData)},
  pages={608--615},
  year={2023},
  organization={IEEE}
}

@misc{Crotti2025AURORAE,
  author    = {Crotti, Ilaria and Cuzzucoli, Alice and DeMarchi, Davide and Ramalli, Edoardo and Selmi, Luigi and Trandafir, Ionut and Pasini, Antonello and Dobricic, Srdjan},
  title     = {AURORAE service: Model description and forecast delivery for the local atmospheric pollution forecasts},
  year      = {2025},
  publisher = {Zenodo},
  doi       = {10.5281/zenodo.15864354},
  url       = {https://doi.org/10.5281/zenodo.15864354}
}

@article{zhu2023machine,
  title={Machine learning in environmental research: common pitfalls and best practices},
  author={Zhu, Jun-Jie and Yang, Meiqi and Ren, Zhiyong Jason},
  journal={Environmental Science \& Technology},
  volume={57},
  number={46},
  pages={17671--17689},
  year={2023},
  publisher={ACS Publications}
}

\end{document}